\documentclass[letterpaper]{article} 
\usepackage{aaai23}  
\usepackage{times}  
\usepackage{helvet}  
\usepackage{courier}  
\usepackage[hyphens]{url}  
\usepackage{graphicx} 
\usepackage{natbib}  
\usepackage{caption} 
\usepackage{algorithm}
\usepackage{algorithmic}

\usepackage{newfloat}
\usepackage{listings}
\DeclareCaptionStyle{ruled}{labelfont=normalfont,labelsep=colon,strut=off} 
\floatstyle{ruled}
\newfloat{listing}{tb}{lst}{}
\floatname{listing}{Listing}
\usepackage[nolist]{acronym}

\begin{acronym}
    \acro{dss}[DSS]{Decision Support System}
    \acroplural{dss}[DSSs]{Decision Support Systems}
    \acro{hds}[HDS]{Human Digital Shadow}
    \acro{iop}[IoP]{Internet of Production}
    \acro{cnn}[CNN]{Convolutional Neural Network}
    \acroplural{cnn}[CNNs]{Convolutional Neural Networks}
    \acro{ds}[DS]{Digital Shadow} 
    \acroplural{ds}[DSs]{Digital Shadows}
    \acro{dt}[DT]{Digital Twin} 
    \acroplural{dt}[DTs]{Digital Twins}
    \acro{ml}[ML]{Machine Learning} 
    \acro{bt}[BT]{Behaviour Tree}
    \acro{bts}[BTs]{Behaviour Trees}
    \acro{htns}[HTNs]{Hierarchical Task Networks}
    \acro{htn}[HTN]{Hierarchical Task Network}
    \acro{fsm}[FSM]{Finite State Machine}
    \acro{fsms}[FSMs]{Finite State Machines}
    \acro{plc}[PLC]{Product Life Cycle}
    \acro{dcc}[DCC]{Digital Capability Centre} 
    \acro{bol}[BOL]{Beginning of Life}
    \acro{ai}[AI]{Artificial Intelligence}
    \acro{xai}[XAI]{eXplainable Artificial Intelligence}
    \acro{hri}[HRI]{Human-Robot Interaction}
    \acro{hrc}[HRC]{Human-Robot Collaboration}
    \acro{iot}[IoT]{Internet of Things}
    \acro{iiot}[IIoT]{Industrial Internet of Things}
    \acro{wwl}[WWL]{World Wide Lab}
    \acro{frp}[FRP]{Fibre Reinforced Plastics} 
    \acro{sus}[SUS]{System Usability Scale}
    \acro{tfd}[TFD]{Temporal Fast Downward}
    \acro{pddl}[PDDL]{Planning Domain Definition Language}
    \acro{lmp}[LMP]{Laser Material Processing}
    \acro{ir}[IR]{Industrial Robot}
    \acroplural{ir}[IRs]{Industrial Robots}
    \acro{tcp}[TCP]{Tool Centre Point} 
    \acro{el}[EL]{Euler Lagrange}
    \acro{rnn}[RNN]{Recurrent Neural Network}
    \acro{rl}[RL]{Reinforcement Learning}
    \acro{am}[AM]{Additive Manufacturing}
    \acro{lpbf}[LPBF]{Laser Powder Bed Fusion}
    \acro{ae}[AE]{Auto Encoder}
    \acro{dl}[DL]{Deep Learning}
    \acro{cpps}[CPPS]{Cyber-Physical Production System}
    \acro{mtl}[MTL]{Metric Temporal Logic}
    \acro{sme}[SME]{Small and Medium sized Enterprise}
    \acroplural{sme}[SMEs]{Small and Medium sized Enterprises}
    \acro{ocel}[OCELs]{Object-Centric Event Log}
    \acroplural{ocel}[OCELs]{Object-Centric Event Logs}
    \acro{gan}[GAN]{Generative Adversarial Network}
    \acroplural{gan}[GANs]{Generative Adversarial Networks}
    \acro{ann}[ANN]{Artificial Neural Network}
    \acroplural{ann}[ANNs]{Artificial Neural Networks}
    \acro{ot}[OT]{Optical Tomography}
    \acro{kpi}[KPI]{Key Performance Indicator}
    \acroplural{kpi}[KPIs]{Key Performance Indicators}
    \acro{ocpc}[OCPC]{Object-Centric Process Cube}
    \acroplural{ocpc}[OCPCs]{Object-Centric Process Cubes}
    \acro{lfd}[LfD]{Learning from Demonstration}
    \acro{mpc}[MPC]{Model Predictive Control} 
    \acro{mip}[MIP]{Mixed Initiative Planning}
    \acro{npc}[NPC]{Non-Player Character}
    \acroplural{npc}[NPCs]{Non-Player Characters}
    \acro{fem}[FEM]{Finite Element Method} 
    \acroplural{fem}[FEMs]{Finite Element Methods}
    \acro{kb}[KB]{Knowledge Base}
    \acroplural{kb}[KBs]{Knowledge Bases}
    \acro{ROS}[ROS]{Robot Operating System}
    \acro{MAS}[MAS]{Multi-Agent System}
    \acroplural{MAS}[MAS]{Multi-Agent Systems}
    \acro{MRS}[MRS]{Multi-Robot System}
    \acroplural{MRS}[MRS]{Multi-Robot Systems}
    \acro{lidar}[LiDAR]{Light Detection And Ranging}
    \acro{slam}[SLAM]{Simultaneous Localisation And Mapping}
    \acro{iqa}[IQA]{Information Quality Assessment}
    \acro{hds}[HDS]{Human Digital Shadow}
    \acro{agv}[AGV]{Automated Guided Vehicle}
    \acro{api}[API]{Abstract Programming Interface}
    \acroplural{api}[APIs]{Abstract Programming Interfaces}
    \acro{w3c}[W3C]{World Wide Web Consortium}    
    \acro{cbf}[CBF]{Control Barrier Function}
    \acroplural{cbf}[CBFs]{Control Barrier Functions}
    \acro{cbfbt}[CBF-BT]{Control Barrier Function Behavior Tree}
    \acroplural{cbfbt}[CBF-BTs]{Control Barrier Function Behavior Trees}
    \acro{care}[CARE]{Constraint-Based Activity REpresentation} 
    \acro{wrt}[w.r.t.]{With Respect To} 
\end{acronym}

\title{Digital Shadows of Safety for Human Robot Collaboration in the World-Wide Lab}
\author {
    Mohamed Behery, 
    Gerhard Lakemeyer
}
\affiliations {
     Knowledge-Based Systems Group (KBSG)\\ RWTH Aachen University\\
    \{behery,gerhard\}@kbsg.rwth-aachen.de
}

\usepackage{bibentry}

\begin{document}

\maketitle

\begin{abstract}
The \ac{wwl} connects the \acp{ds} of processes, products, companies, and other entities allowing the exchange of information across company boundaries.
Since \acp{ds} are context- and purpose-specific representations of a process, as opposed to \acp{dt} which offer a full simulation, the integration of a process into the \ac{wwl} requires the creation of \acp{ds} representing different aspects of the process.
\ac{hrc} for assembly processes was recently studied in the context of the \ac{wwl} where \acp{bt} were proposed as a standard task-level representation of these processes.
We extend previous work by proposing to standardise safety functions that can be directly integrated into these \acp{bt}.
This addition uses the \ac{wwl} as a communication and information exchange platform allowing industrial and academic practitioners to exchange, reuse, and experiment with different safety requirements and solutions in the \ac{wwl}.
\end{abstract}

\section{Introduction}
Industrial robotics processes, particularly those involving \ac{hri}, have recently become widespread in many manufacturing domains and interaction paradigms~\cite{baier2022framework}.
Despite being dynamic and safety critical applications of robotics, these processes usually involve ad hoc programming which is often verified empirically on a case-by-case basis, especially when the process relies on \ac{dl} applications.
In this paper, we propose standardizing safety functions for \ac{hri} processes in the \ac{wwl}.
Having a standard representation offers the ability to reuse these functions as well as producing reusable code for representing the tasks involved in a process.

Previous work has introduced the \acf{ds} of production along with requirements and challenges facing a standard \ac{ds}~\cite{bauernhansl2018digital}.
Further research with a focus on integrating \ac{hri} assembly into the \ac{wwl} suggested using \acfp{bt}~\cite{colledanchise2018behavior} as a standard task-level representation for these processes~\cite{behery2023digital}.

\acp{bt} model a robot's behavior in a tree structure where the leaves represent atomic actions and conditions while internal nodes are used for control flow. 
Execution begins by \textit{ticking} the root which propagates this \textit{tick} down to the leaves.
When a node is ticked, it returns its execution status to its parent allowing real-time execution monitoring as well as plan synthesis and repair.

This paper proposes using standard parameterisable safety functions encapsulated in \ac{bt} nodes.
We propose using \acp{cbf}~\cite{ames2019control} which have been combined with traditional \acp{bt}~\cite{ozkahraman2020combining} producing \acp{cbfbt} that guarantee convergence and safety.

The rest of the paper is structured as follows:
The next section~\ref{sec:background} describes the background knowledge and related work on \acp{ds} in the \ac{wwl}, \acp{bt}, and \acp{cbf}.
After that we discuss our proposal for standard \acp{cbf} to represent different safety aspects in \ac{hri} processes.
Lastly, we conclude the paper with a summary and recommendations for future work.

\begin{figure*}[t]
    \centering
    \includegraphics[width=0.75\textwidth]{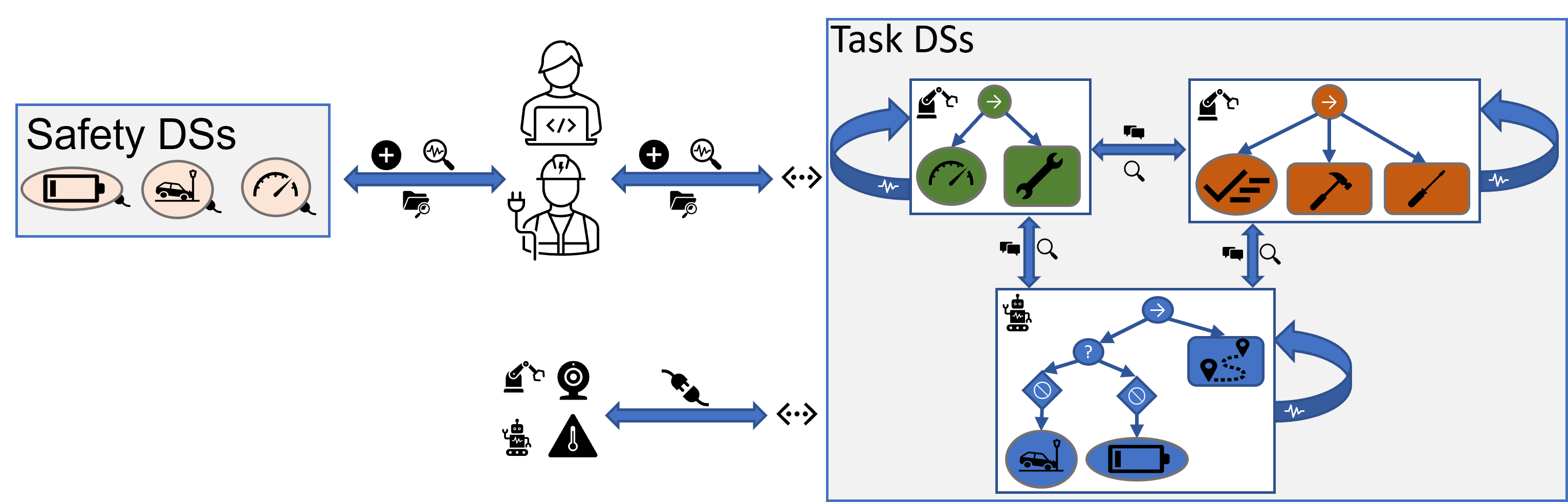}
    \caption{The architecture proposed by~\citet{behery2023digital} integrated into the \ac{wwl}~\cite{liebenberg2020information}. 
    Human operators add, introspect, and query \acp{ds} of task representations of different agents.
    They can also share and use condition nodes encapsulating a \ac{cbf} to an existing \ac{bt}.
    Additionally, \acp{ds} can communicate with each other and self-optimize as proposed by~\cite{bauernhansl2018digital}.}
    \label{fig:wwl-overview}
\end{figure*}

\section{Background and Related Work}
\label{sec:background}
The \ac{wwl} provides an infrastructure for communication between different entities within a company as well as beyond its boarders~\cite{liebenberg2020information}.
Entities in the \ac{wwl} are represented using their \acp{ds} which are task- and purpose-specific projections of entities such as processes, machines, and products to name a few.    
Compared to the widely used \acp{dt}, which represent a full simulation of an entity~\cite{liebenberg2020information}, \acp{ds} can be used in real-time applications such as quality prediction and plan repair~\cite{liebenberg2021autonomous}.
Previous work~\cite{bauernhansl2018digital} defined requirements for \acp{ds} split into different levels of complexity and development milestones.

Since robotics and particularly \ac{hrc}-based processes span different disciplines: motor control, perception, planning, and human safety among others, they require multiple \acp{ds} to be fully integrated into the \ac{wwl}.
Task level representations of different robotic activities have been in development over the past several decades.
They usually involve modeling the domain, e.g., \ac{pddl}~\cite{mcdermott2003formal} where a developer has to model the robot's capabilities as well as the initial and goal states of the world.
After that the agent can create a plan to get from the former to the latter.
Recent approaches have shown combinations and transformations between these systems and rule based systems~\cite{niemueller2019goal} that combines goal reasoning and teleoreactive programs.
However, these paradigms are not modular enough to allow code reuse and exchange due to their dependence on problem, domain, and robot descriptions.

\acp{bt}~\cite{colledanchise2018behavior} were recently adapted by the robotics community to overcome these problems.
Their structure allows entire tree branches to be changed or removed without affecting the rest of tree.
Modifying the tree requires less effort and time compared to other graph-based models such as \acp{fsm} or \acp{htn}, since the transitions are encoded in the tree structure and do not have to be explicitly modeled and maintained with every change.

They were recently combined with \acp{cbf} to guarantee their \textit{concurrent} goal satisfaction allowing a tree to work toward multiple given goals without breaking safety rules~\cite{ozkahraman2020combining}.
A \ac{cbf} is a function $h:R^{n}\rightarrow R$ is used to evaluate the system state $x$ \acs{wrt} safety where $h(x) \ge 0$ implies $x$ is a safe state.  
They were encapsulated in condition nodes of the \ac{bt} which can then select appropriate actions to keep the robot within a safe set of states without contradicting previously satisfied conditions.

\section{\acp{cbf} with Predefined \acsp{api} as Safety \acp{ds}}
\label{sec:contrib}

\citet{ozkahraman2020combining} combined \acp{cbf} with \acp{bt} to achieve goals with different priorities.
Goal conflicts are resolved using the tree's structure to allow the robots to favor safety over task execution.
We propose maintaining a pool of condition nodes encapsuating \acp{cbf} through the \ac{wwl} allowing practitioners from academia and industry to share and exchange standard and explicit safety requirements across company boundaries.
Exchanging process \acp{ds} through the \ac{wwl} has been discussed and demonstrated in previous work~\cite{liebenberg2021autonomous} where process \ac{ds} were used for real-time plan repair in several use-cases.

\citet{behery2023digital} proposed standardizing \acp{bt} as a task-level \ac{ds} for assembly processes involving \ac{hri}.
The modularity of \acp{bt} allows nodes, branches, and entire sub-trees to be added, removed, and exchanged thus adding new levels of safety (e.g., guarding actions with preconditions) or alternative behaviors in case of failure.

Our extension of the \ac{wwl} architecture proposed in~\cite{liebenberg2021autonomous,behery2023digital} is shown in Figure~\ref{fig:wwl-overview}.
In addition to storing, sharing, and querying task level description, we allow human operators to exchange safety functions such as checking battery, joint velocities, or whether a human is in the cell of the robot.
Each of these functions can have it's own interface to the middle-ware (e.g., ROS) of a robot allowing it to read battery or joint state or other perception information such as 3D point cloud data.
As a parameter, they can take topic names of these types of inputs and connect to them directly, since they are condition nodes, their return status encoding whether the system is in a safe state.
Since \acp{cbf} are composable, they can form safety branches and be added to a \ac{bt} for added safety without affecting the core tasks of the tree.

\section{Conclusion and Future Work}
\label{sec:conc}
We propose using standard \acp{ds} for safety in \ac{hrc}-based assembly processes.
Storing \acp{cbf} encapsulated in \ac{bt} condition nodes as proposed by~\cite{ozkahraman2020combining} allows exchanging and reusing them within the \ac{wwl}.
These functions can be used in benchmarks for evaluating the safety of newly proposed approaches in different paradigms of \ac{hrc}.
Some topics still remain open such as the creation of a general \ac{api} between the \ac{cbf} and the perception modules or the underlying knowledge-bases.
Another direction is the interface for defining these functions, their parameters, particularly involving training requiring generalization guarantees.
This aligns with the road map presented by ~\citet{behery2023digital} allowing automatic registration, modification, and optimization of \acp{ds} in the \ac{wwl}.

\section*{Acknowledgements}
Funded by the Deutsche Forschungsgemeinschaft (DFG, German Research Foundation) under Germany's Excellence Strategy -- EXC-2023 Internet of Production -- 390621612. This work is supported by the EU ICT-48 2020 project TAILOR (No. 952215).

\bibliography{main}

\end{document}